\documentclass[letterpaper, 10 pt, conference]{ieeeconf}  
\usepackage{graphicx}
\usepackage{amssymb}
\usepackage{amsmath}
\usepackage{booktabs}
\usepackage{tabularx}
\usepackage{supertabular}
\usepackage{threeparttable}
\usepackage{color}
\usepackage{multirow}
\usepackage{cite}
\usepackage{makecell}
\usepackage{bm}
\usepackage{pifont}
\makeatletter
\let\NAT@parse\undefined
\makeatother
\usepackage[colorlinks=true,
citecolor=blue,
linkcolor=blue,
anchorcolor=blue,
urlcolor=blue]{hyperref}
\UseRawInputEncoding

\IEEEoverridecommandlockouts                              

\title{\LARGE \bf
TDFANet: Encoding Sequential 4D Radar Point Clouds Using Trajectory-Guided Deformable Feature Aggregation for Place Recognition
}

\author{Shouyi Lu$^{1}$, Guirong Zhuo$^{1*}$, Haitao Wang$^{2}$, Quan Zhou$^{1}$, Huanyu Zhou$^{1}$, Renbo Huang$^{1}$, \\ Minqing Huang$^{1}$, Lianqing Zheng$^{1}$, and Qiang Shu$^{3}$
	\thanks{$^*$ Corresponding author.} 
	\thanks{This work was supported by the National Natural Science Foundation of China under Grant 52325212, the National Key Research and Development Program of China (No. 2022YFE0117100), and the Fundamental Research Funds for the Central Universities.}
	\thanks{$^{1}$Shouyi Lu, Guirong Zhuo, Quan Zhou, Huanyu Zhou, Renbo Huang, Minqing Huang, and Lianqing Zheng are with the School of Automotive Studies, Tongji University, Shanghai, China.}%
	\thanks{$^{2}$Haitao Wang is with the Shanghai Geometrical Perception and Learning Co., Ltd., Shanghai, China.}
	\thanks{$^{3}$Qiang Shu is with the Shanghai Tongyu Automotive Technology Co., Ltd., Shanghai, China}%
}

\begin{document}

\maketitle
\thispagestyle{empty}
\pagestyle{empty}

\begin{abstract}
Place recognition is essential for achieving closed-loop or global positioning in autonomous vehicles and mobile robots.
Despite recent advancements in place recognition using 2D cameras or 3D LiDAR, it remains to be seen how to use 4D radar for place recognition - an increasingly popular sensor for its robustness against adverse weather and lighting conditions. 
Compared to LiDAR point clouds, radar data are drastically sparser, noisier and in much lower resolution, which hampers their ability to effectively represent scenes, posing significant challenges for 4D radar-based place recognition.
This work addresses these challenges by leveraging multi-modal information from sequential 4D radar scans and effectively extracting and aggregating spatio-temporal features.
Our approach follows a principled pipeline that comprises (1) dynamic points removal and ego-velocity estimation from velocity property, (2) bird's eye view (BEV) feature encoding on the refined point cloud, (3) feature alignment using BEV feature map motion trajectory calculated by ego-velocity, (4) multi-scale spatio-temporal features of the aligned BEV feature maps are extracted and aggregated.
Real-world experimental results validate the feasibility of the proposed method and demonstrate its robustness in handling dynamic environments.
Source codes are available.
\end{abstract}

\section{INTRODUCTION}
Place recognition is a fundamental task in mobile robotics, widely applied in scene understanding, loop closure detection, and localization. State-of-the-art techniques in place recognition primarily use LiDAR \cite{ptcnet,ndttransformer,cvtnet,overlaptransformer}, camera \cite{cricavpr,patchvlad,transvpr}, or their fusion \cite{lcpr,adafusion,minkloc++}. Surprisingly, the potential of 4D radars remains under-explored. As an emerging sensor, 4D radar is gaining attention due to its improved imaging ability, robustness in adverse weather conditions, ability to measure object velocities, and cost-effectiveness \cite{ratrack,rcfusion,hidden,10611470}. These advantages make 4D radar a compelling and robust alternative to LiDAR.

Despite the advantages, 4D radar produces sparse point cloud scans with lower spatial resolution than 3D LiDAR. A single-frame 4D radar point cloud typically contains only a few hundred points, roughly 1\% of the number captured by 32-beam LiDAR. Besides, aliasing caused by multipath echoes may blur the shape of static architectures and dynamic objects. These factors diminish the precision of the 4D radar scene depiction. Therefore, how to address the issues of sparsity and noise in 4D radar point clouds and enhance their scene representation capabilities is the major challenge of applying 4D radar to the place recognition task.

To address these issues and unleash the potential of 4D radar, we propose a novel 4D radar place recognition model, termed TDFANet, designed to encode sequential 4D radar point clouds. TDFANet encodes a finite-length sequence of 4D radar point clouds, exploiting the spatio-temporal context information present in a sequence of 4D radar point sets to enhance the scene representational capacity of the sparse point clouds. Specifically, RANSAC-based ego-velocity estimation strategy is first employed to remove dynamic points and estimate radar ego-velocity. Then after preprocessing, the refined point clouds are encoded into bird's eye view (BEV) feature maps. To mitigate the feature shift between multiple BEV feature maps resulting from the vehicle's 3D motion, we propose a trajectory-guided feature alignment module that aligns multiple BEV feature maps using motion trajectories estimated from ego-velocity, ensuring accurate aggregation of features for the same object. We then propose a novel spatio-temporal pyramid deformable feature aggregation module, which constructs a spatio-temporal pyramid architecture with deformable attention \cite{deformable} to extract and aggregate multi-scale spatio-temporal features from sequential BEV feature maps. Finally, the aggregated features are compressed using Generalized-Mean pooling (GeM) \cite{gem} to generate the global descriptor. For evaluation, we collected a multi-modal dataset using a vehicle equipped with 4D radar and other sensors. Real-world evaluation results demonstrate the feasibility of TDFANet. Overall, the contributions of this work are summarized as follows:
\begin{itemize}
\item{A novel encoding architecture, TDFANet, is proposed. It is the first end-to-end network to employ sequential 4D radar scans for place recognition.}	
\item{A trajectory-guided feature alignment method, leveraging the velocity property of 4D radar scan, is proposed to align multiple BEV feature maps.}	
\item{A spatio-temporal pyramid deformable feature aggregation method is proposed to extract and aggregate multi-scale spatio-temporal features of sequential BEV feature maps.}	
\item{A dataset is constructed for 4DRPR, on which our TDFANet is validated.}	
\end{itemize}
\section{RELATED WORK}

This work focuses on place recognition techniques based on point clouds, which can be divided into two categories depending on the sensor modality: 3D LiDAR Place Recognition (3DLPR) and 4D Radar Place Recognition (4DRPR).

\textbf{3D LiDAR Place Recognition (3DLPR)}. ScanContext \cite{scancontext} describes scenes by projecting 3D point clouds onto a two-dimensional plane and segmenting them to generate context histograms. As neural network technology advances, data-driven descriptors have increasingly gained popularity. PointNetVLAD \cite{pointnetvlad} is the seminal end-to-end architecture for 3DLPR. PPT-Net \cite{pptnet} proposes a pyramid point transformer module that adaptively learns spatial relationships among neighboring points within the point cloud. MinkLoc3D \cite{minkloc3d} utilizes sparse 3D convolutions to generate  point cloud descriptors.

The above methods rely on single-frame observations. Despite achieving reasonable results, they are less reliable when used for long-time-span place recognition with appearance changes. To address this limitation, several studies have employed sequential LiDAR scans to improve the performance of long-term place recognition. SeqLPD \cite{seqlpd} proposes a coarse-to-fine sequence matching approach based on LPDNet \cite{lpdnet}. SeqOT \cite{seqot} combines the Transformer model \cite{transformer} and GeM pooling \cite{gem} to aggregate multiple LiDAR scans into a global descriptor. These sequence-based strategies have notably enhanced performance in complex environments, especially in scenarios characterized by long-term appearance changes and dynamic disturbances. \textit{In line with this trend, we propose a novel model for place recognition using sequential 4D radar point clouds, which leverages the unique properties of 4D radar to extract and aggregate spatio-temporal features of sequential radar scans}.

\textbf{4D Radar Place Recognition (4DRPR).} Among early attempts at radar place recognition, Kidnapped Radar \cite{kidnapped} provides a rotation-invariant solution for spinning radars. AutoPlace \cite{autoplace} utilizes sequential automotive radars and employs an LSTM network \cite{lstm} for spatio-temporal feature embedding. With the advent of 4D radar, 4DRPR has garnered increasing attention in the academic community. NTU4DRadLM \cite{ntu4dradlm} proposes a loop closure detection module based on Intensity Scan Context \cite{intensity}. TransLoc4D \cite{transloc4d} combines sparse convolution and Transformer to generate 4D radar descriptors. \textit{Our TDFANet is the first place recognition method to leverage sequential 4D radar point clouds, enhancing scene representation by effectively integrating sequential 4D radar spatio-temporal information}.
\section{METHOD}
\label{Method}
\begin{figure*}[tb]
	\centering
	\includegraphics[width=0.95\linewidth]{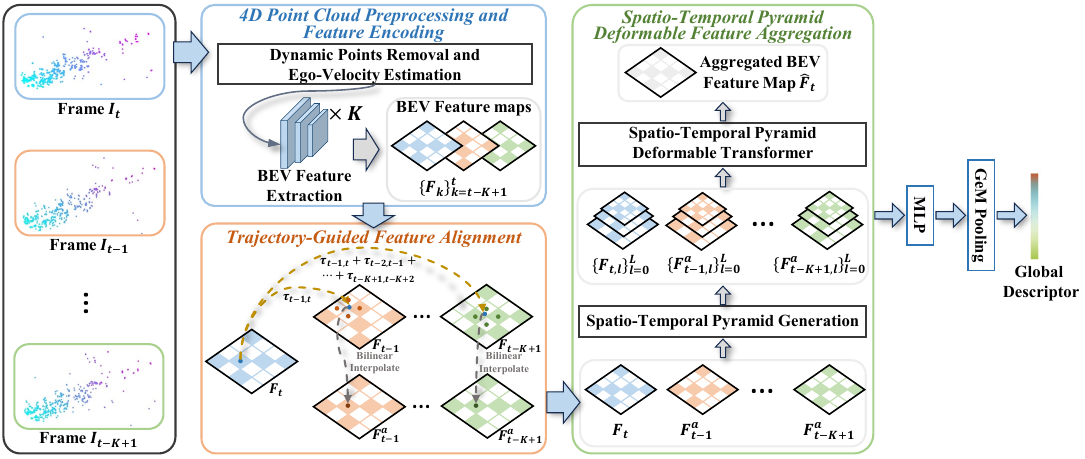}
	\centering
	\caption{TDFANet Overview: Given sequential 4D radar point clouds. First, preprocessing is performed to refine the 4D radar point cloud based on ego-velocity regression and RANSAC filtering. The refined point clouds are then encoded into BEV feature maps. Next, a trajectory-guided feature alignment method is proposed to align BEV feature maps at different time steps. Subsequently, a spatio-temporal pyramid deformable feature aggregation method is proposed to aggregate the aligned BEV feature maps. Finally, the final global descriptor is generated using GeM pooling.}
	\label{fig2}
\end{figure*}
This section expounds on the details of the TDFANet. The inputs to the network are the latest $K$ 4D radar point clouds $\{I_k \in \mathbb{R}^{N \times 5}\}_{k=t-K+1}^t$, where each 4D radar scan $I_k$ includes the point cloud Cartesian coordinates $P_k=\{\bm{p}_i|\bm{p}_i\in\mathbb{R}^{3}\}_{i=1}^N$, radial velocity $V_k=\{v^d_i|v^d_i\in\mathbb{R}^{1}\}_{i=1}^N$, and radar cross section $RCS_k=\{rcs_i|rcs_i\in\mathbb{R}^{1}\}_{i=1}^N$, where $N$ denotes the numbers of points. These  point clouds first undergo preprocessing and BEV feature encoding as introduced in Sec. \ref{PCP}. To correct feature misalignment between different BEV feature maps caused by ego-vehicle motion, Sec. \ref{TGFA} introduces a trajectory-guided feature alignment method. Subsequently, Sec. \ref{STPDFA} presents a spatio-temporal pyramid deformable feature aggregation method for comprehensive feature aggregation across multiple spatial and temporal scales. Finally, Sec. \ref{DG} introduces GeM pooling to compress the aggregated BEV feature maps and generate a 1-D global descriptor. Fig. \ref{fig2} illustrates the overall TDFANet framework.

\subsection{4D Point Cloud Preprocessing and Feature Encoding}
\label{PCP}
\subsubsection{Dynamic Points Removal and Ego-velocity Estimation} Outdoor scenes typically contain numerous dynamic objects, such as vehicles and pedestrians, which do not maintain temporal consistency. When revisiting the same place, these moving objects may no longer be present, potentially leading to errors in place recognition. To address this challenge, the velocity property of 4D radar is employed to filter out dynamic points and estimate the radar ego-velocity.

Given a 4D radar point cloud with $N$ points $\{\bm{p}_i, v^d_i, rcs_i\}_{i=1}^N$, the radial velocity $v^d_i$ of each point is formally the product of its velocity relative to the radar $\bm{v}_i^r=[v_{x,i}^r,v_{y,i}^r,v_{z,i}^r]^T$ and the unit vector of the direction relative to the radar $\bm{\hat{p}}_i=\frac{\bm{p}_i}{\|\bm{p}_i\|}=[\hat{p}_{x,i},\hat{p}_{y,i},\hat{p}_{z,i}]^T$:
\begin{equation}
	\label{eq1}
	v^d_i=\bm{\hat{p}}_i^T\bm{v}_i^r=\hat{p}_{x,i}v_{x,i}^r+\hat{p}_{y,i}v_{y,i}^r+\hat{p}_{z,i}v_{z,i}^r.
\end{equation}

All static points in the point cloud share the same velocity relative to the radar, which is equal in magnitude but opposite in direction to the ego-velocity of the 4D radar, i.e., $\bm{v}^r=-\bm{v}^e=[-v_x^e,-v_y^e,-v_z^e]^T$. Given $N'$ static points, applying Eq. \ref{eq1} to each point yields the following system of linear equations:
\begin{equation}
	\label{eq2}
	\begin{bmatrix}
		v_1^d \\
		\vdots\\
		v_{N'}^d
	\end{bmatrix}
=
\begin{bmatrix}
	\hat{p}_{x,1} & \hat{p}_{y,1} & \hat{p}_{z,1}\\
	\vdots & \vdots & \vdots\\
	\hat{p}_{x,N'} & \hat{p}_{y,N'} & \hat{p}_{z,N'}
\end{bmatrix}
\begin{bmatrix}
	-v_x^e\\
	-v_y^e\\
	-v_z^e
\end{bmatrix}.
\end{equation}

It can be expressed as: $\bm{v}^d=-\hat{\bm{P}}\bm{v}^e$. When the matrix $\hat{\bm{P}} \in \mathbb{R}^{N' \times 3}$ is full rank, the ego-velocity $\bm{v}^e$ can be determined by solving Eq. \ref{eq2} using the normal equation method \cite{numerical}:
\begin{equation}
	\label{eq3}
	\bm{v}^e=-(\hat{\bm{P}}^T\hat{\bm{P}})^{-1}\hat{\bm{P}}^T\bm{v}^d.
\end{equation}

However, dynamic points in the 4D radar point cloud can disrupt the accurate estimation of ego-velocity. To address this, we employ the Random Sample Consensus (RANSAC) method \cite{ransac}, treating dynamic points as outliers and static points as inliers. By iteratively sampling, we estimate the ego-velocity $\bm{v}^e$ and identify outliers to detect dynamic points. By this means, the velocity property of the 4D radar scan is leveraged to regress the radar ego-velocity $\bm{v}^e$, while RANSAC is applied to filter out dynamic points from the point cloud.

\subsubsection{BEV Feature Encoding} After obtaining the 4D radar point cloud with dynamic points removed, we exploit the pillar \cite{pointpillars} representation to encode 4D radar features, which directly converts the 4D radar input to a pseudo image in the bird's eye view (BEV). Then we extract 4D radar features with a pillar feature network, obtaining a 4D radar BEV feature map $F \in \mathbb{R}^{C\times H\times W}$.

After processing each frame in the sequential 4D radar point cloud $\{I_k \in \mathbb{R}^{N \times 5}\}_{k=t-K+1}^t$ following the steps described above, we obtain $K$ radar ego-velocities $\{\bm{v}^e_k\}_{k=t-K+1}^t$ and BEV feature maps $\{F_k\}^t_{k=t-K+1}$.

\subsection{Trajectory-Guided Feature Alignment}
\label{TGFA}
The vehicle's 3D motion results in feature shifts in the multi-frame BEV feature maps, reducing the effectiveness of feature aggregation. Inspired by the insight that ego-velocity-based trajectory estimates can provide valuable clues for feature alignment, this work proposes a trajectory-guided feature alignment method (TGFA) to align the sequential BEV feature maps $\{F_k\}^t_{k=t-K+1}$.

Specifically, we assume the vehicle moves at a constant speed between consecutive 4D radar scans. Based on the estimated radar ego-velocity $\{\bm{v}^e_k\}_{k=t-K+1}^t$, the radar's displacement in the $X$ direction $\{s^x_{k-1,k}\}_{k=t-K+2}^t$ and $Y$ direction $\{s^y_{k-1,k}\}_{k=t-K+2}^t$ between consecutive radar scans can be calculated:
\begin{equation}
	\label{eq4}
	\begin{aligned}
		&s^x_{k-1,k}=v^e_{x,k-1} / f_r,\\
		&s^y_{k-1,k}=v^e_{y,k-1} / f_r,\\
	\end{aligned}
\end{equation}
where $f_r$ denotes the frame rate of the 4D radar. Given the radar displacement and the grid size in the BEV feature map, the change in grid coordinates can be calculated:
\begin{equation}
	\label{eq5}
	\begin{aligned}
		&\Delta x_{k-1,k}=s^x_{k-1,k}/h,\\
		&\Delta y_{k-1,k}=s^y_{k-1,k}/w,\\
	\end{aligned}
\end{equation}
where $h$ and $w$ represent the height and width of the grid, respectively. Therefore, the grid motion trajectory $\{\tau_{k-1,k} = (\Delta x_{k-1,k}, \Delta y_{k-1,k})\}_{k=t-K+2}^{t}$ between consecutive BEV feature maps can be established, representing the coordinate changes over time. As shown in Fig. \ref{fig2}, this motion trajectory allows us to determine the position of each grid in $F_t$ within $\{F_k \in \mathbb{R}^{C \times H \times W}\}_{k=t-K+1}^{t-1}$. Given that the position is usually represented as a floating-point number, we employ bilinear interpolation to retrieve the corresponding features, thus producing the aligned BEV feature map $\{F_k^a \in \mathbb{R}^{C \times H \times W}\}_{k=t-K+1}^{t-1}
$.
\begin{figure*}[tb]
	\centering
	\includegraphics[width=0.95\linewidth]{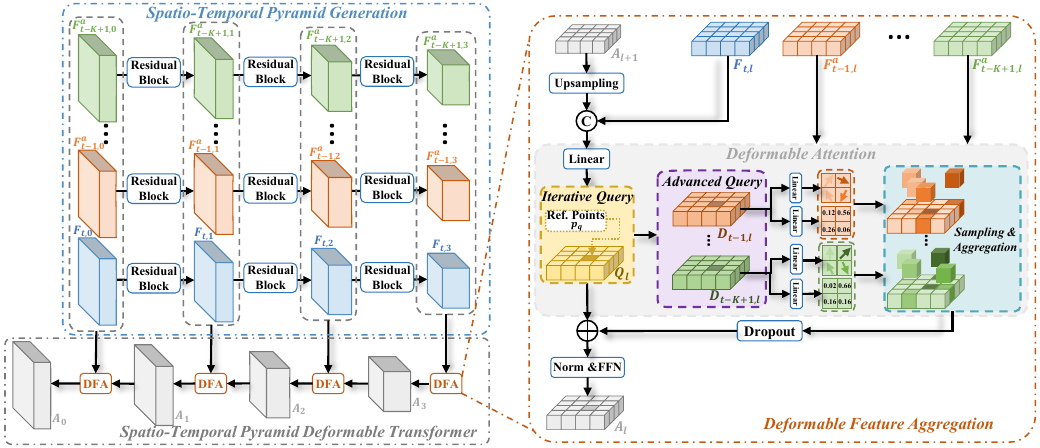}
	\centering
	\caption{Spatio-Temporal Pyramid Deformable Feature Aggregation: A spatio-temporal feature pyramid is built using residual blocks, followed by the introduction of a spatio-temporal pyramid deformable transformer to aggregate these features. We illustrate the deformable feature aggregation process using the reference point $p_q$ as an example.}
	\label{fig4}
\end{figure*}

\subsection{Spatio-Temporal Pyramid Deformable Feature Aggregation}
\label{STPDFA}
Due to the sparseness, unevenness, and disorder of point clouds, it is challenging to extract robust features. Moreover, TGFA relies on grid motion trajectory derived from radar ego-velocity but ignores rotational effects. To address these issues and achieve comprehensive feature aggregation, we propose a spatio-temporal pyramid deformable feature aggregation method (STPDFA), as shown in Fig. \ref{fig4}. This method comprises two key components: spatio-temporal pyramid generation and spatio-temporal pyramid deformable transformer.

\subsubsection{Spatio-Temporal Pyramid Generation} We encode the input BEV feature maps $F_t,F_{t-1}^{a},...,F_{t-K+1}^{a}$ using three hierarchical layers composed of residual blocks \cite{resnet} to construct a spatio-temporal feature pyramid. Each residual block includes a convolutional layer with a stride of 2 to downsample the features. Ultimately, we obtain multi-scale spatio-temporal features $\{F_{t,l} \in \mathbb{R}^{2^l \times C\times \frac{H}{2^l} \times \frac{W}{2^l}} \}^3_{l=0}$,  $\{F_{t-1,l}^{a}\}^3_{l=0}$,$...$, $\{F_{t-K+1,l}^{a}\}^3_{l=0}$.

\subsubsection{Spatio-Temporal Pyramid Deformable Transformer} We redesign the deformable attention \cite{deformable} and integrate it with the spatio-temporal pyramid, proposing a spatio-temporal pyramid deformable transformer (STPDT). STPDT begins with the bottom-level features of the spatio-temporal pyramid and iteratively aggregates sequential BEV feature maps across different spatial scales. The key to this process is the continuous updating of iterative query features $Q_l$ across the $L$ deformable feature aggregation layers (DFA). $Q_l$ incorporates the aggregated features from the previous layer as spatial priors to guide the current layer’s feature aggregation process, where $l$ represents the index of the deformable feature aggregation layer.

Specifically, in the first deformable feature aggregation layer, $Q_3$ is initialized by $F_{t,3}$. In subsequent deformable feature aggregation layers, $Q_l$ is jointly determined by $F_{t,l}$ and the aggregated features $A_{l+1}$ from the previous layer. First, bilinear interpolation is used to upsample $A_{l+1}$ to match the feature map size of $F_{t,l}$, after which it is concatenated with $F_{t,l}$ along the feature dimension. Finally, a linear transformation is applied for dimensionality reduction. The process for generating and updating the iterative query features $Q_l$ is expressed as follows:
\begin{equation}
	\label{eq6}
	Q_l = 
	\begin{cases}
		F_{t,l} \quad \text{for} \quad l=3\\
		\text{Linear}([F_{t,l},\text{Upsample}(A_{l-1})]) \quad \text{for} \quad l < 3 ,
	\end{cases}
\end{equation}
where $\text{Linear}(\cdot)$ represents the linear transformation, $\text{Upsample}(\cdot)$ represents the upsampling operation, and $[\cdot,\cdot]$ represents channel-wise concatenation.

Due to inherent structural limitations, existing deformable DETR \cite{deformable} cannot adaptively adjust the deformable masks for different BEV feature maps. To address this, we concatenate the iterative query features $Q_l$ with the previous $K$-$1$ aligned BEV feature maps $\{F_{t-k,l}^a\}_{k=K-1}^{1}$, and apply a linear transformation to reduce dimensionality, resulting in advanced query features $\{D_{t-k,l}\}_{k=K-1}^{1}$, which  integrates information from both query frame and target frame thereby generating more precise offsets and weights for the deformable mask applied to different target frame.

\begin{equation}
	\label{eq7}
	D_{t-k,l}=\text{Linear}([Q_l,F_{t-k,l}^a]).
\end{equation}

Next, for the previous $K$-$1$ aligned BEV feature maps, we simultaneously apply the advanced query features $D_{t-k,l}$ to perform deformable attention on the corresponding BEV feature maps $F_{t-k,l}^a$, generating intermediate features $M_{t-k,l}$: 
\begin{equation}
	\label{eq8}
	M_{t-k,l} = \text{DeformableA}(D_{t-k,l}, F_{t-k,l}^a),
\end{equation}
where $\text{DeformableA}(\cdot,\cdot)$ represents the deformable attention operation. Finally, the aggregated feature $A_l$ is calculated as follows:
\begin{equation}
	\label{eq9}
	A_l = \text{FFN}(\text{LN}(\text{Dropout}(\sum_{k=1}^{K-1} M_{t-k,l}) + Q_l)),
\end{equation}
where $\text{FFN}(\cdot)$ denotes feed-forward network \cite{transformer}, $\text{LN}(\cdot)$ denotes the layer normalization \cite{ln}, $\text{Dropout}(\cdot)$ denotes the drop-out operation \cite{dropout}. After $L$ layers of iterative query aggregation, the aggregated BEV feature map can finally be obtained as $\hat{F}_t=A_0$.
\setlength{\tabcolsep}{0.9mm}
\begin{table*}[t]
	\caption{Comparison with SOTA methods for LiDAR/radar-based place recognition on 4D radar datasets. `$^*$' denotes using sequential frames for place recognition. The best result for each sequence is bold, and the second best is underlined.}	
	\centering
	\footnotesize
	\begin{center}
		\resizebox{0.95\textwidth}{!}
		{
			\begin{tabular}{l||ccc|ccc|ccc|ccc|ccc}
				\toprule
				&  \multicolumn{3}{c|}{Seq 1-2}  &\multicolumn{3}{c|}{Seq 2-2}      & \multicolumn{3}{c|}{Seq 3-2} & \multicolumn{3}{c|}{Seq 3-3} &  \multicolumn{3}{c}{Seq All} \\ 
				\cline{2-16}\noalign{\smallskip}
				
				\multirow{-2}{*}{\begin{tabular}[c]{@{}c@{}}Method \end{tabular}}
				&  r@1  & r@5 & r@10 & r@1 & r@5 & r@10  & r@1 & r@5  & r@10 & r@1   & r@5 & r@10 &r@1 &r@5 &r@10 \\
				\hline\hline
				\noalign{\smallskip}
				
				ScanContext \cite{scancontext} 
				&76.01 &94.93 
				&96.88 &86.88 
				&\underline{95.25} &96.18 
				&95.13 &98.79 
				&98.84 &70.52 
				&85.97 &89.83 
				&78.19 &90.10 
				&91.75 
				\\
				MinkLoc3D \cite{minkloc3d}
				&94.90	&98.76	
				&99.53	&85.45	
				&93.26	&95.42	
				&92.49	&97.73	
				&99.08  &61.12	
				&73.51	&78.79  
				&83.81  &90.56 
				&92.58 
				\\
				PPT-Net \cite{pptnet}
				&91.63	&96.88	
				&97.62	&82.36	
				&91.32	&94.21	
				&91.28	&97.83
				&98.89	&64.21			
				&78.95  &82.87	
				&81.59	&89.39	
				&91.70
				\\
				\hline 
				\noalign{\smallskip} 
				Autoplace$^*$ \cite{autoplace}  
				&\underline{97.62}	&98.83	
				&99.14	&\underline{91.41}	
				&95.06	&\underline{96.35}	
				&\underline{97.83}	&\underline{99.57}	
				&\underline{99.86}	&\underline{79.17}	
				&\underline{89.94}	&\underline{93.53}	
				&\underline{91.15}	&\underline{94.91}			
				&\underline{96.39}	
				\\ 
				SeqOT$^*$ \cite{seqot}    
				&97.08	&\underline{99.13}	
				&\bf{99.97}	&83.99	
				&90.61	&93.97	    
				&96.75	&99.22	
				&99.76	&71.98	
				&83.77	&90.12	 
				&87.40  &92.44	
				&95.23
				\\           					
				Ours      
				& \bf{99.63}	& \bf{99.87}	
				& \bf{99.97}	& \bf{97.33}	
				& \bf{99.19}	& \bf{99.44}	
				& \bf{99.52}	& \bf{99.62}	 
				& \bf{99.90}	& \bf{83.36}	  
				& \bf{92.17}	& \bf{94.89}	   
				& \bf{95.01}	& \bf{97.40} 
				& \bf{98.18}   
				\\ \bottomrule
			\end{tabular}
		}
	\end{center}
	\label{table:sota}
\end{table*}
\subsection{Descriptor Generation and Network Training}
\label{DG}
To generate the final 4D radar point cloud descriptor $U\in \mathbb{R}^{256}$, we first map the channels of the aggregated BEV feature map $\hat{F}_t$ to 256 dimensions using a multilayer perceptron. Then, we apply GeM pooling \cite{gem} to compress the feature map into a compact global descriptor vector.

To train our TDFANet network, we adopt the Lazy Quadruplet loss function \cite{pointnetvlad}, which is defined as follows:
\begin{equation}
	\label{eq12}
	\begin{aligned}
		L_Q =& \max \limits_{j}([\alpha+d_E(U_q,U_{pos})-d_E(U_q,U_{neg_j})]_+)+\\&[\beta+d_E(U_q,U_{pos})-d_E(U_q,U_{neg^*})]_+,
	\end{aligned}
\end{equation}
where $\alpha$ and $\beta$ denote the predefined margin, $d_E(\cdot,\cdot)$ represents Euclidean distance, $U_q$ represents the descriptor corresponding to the query sample, $U_{pos}$ denotes the descriptor corresponding to the best positive matching sample, $U_{neg_j}$ represents the descriptors corresponding to the true negative samples, $U_{neg^*}$ denotes the descriptor corresponding to the hard negative sample, and $[...]_+$ indicates the hinge loss.

\section{EXPERIMENTAL SETUP}
\subsection{Dataset}
\begin{figure}[t]
	\centering
	\includegraphics[scale=0.9]{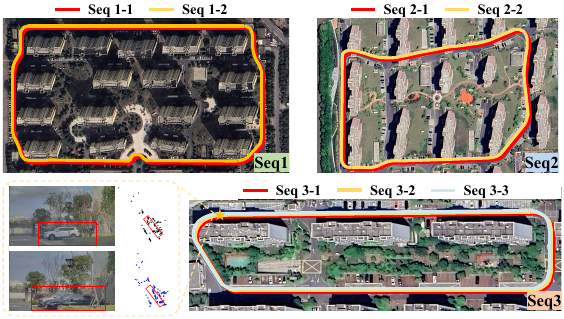}
	\caption{Overview of the dataset we collected. The trajectories in different colors represent data collected during different time periods. Example is given for vegetation and vehicle changes due to the long time span.}
	\label{fig5}
\end{figure}

Given the absence of a publicly available dataset specifically designed for 4DRPR, we utilize a vehicle equipped with a 4D radar, cameras, and RTK GPS as our data collection platform. Data are gathered from seven sequences across three distinct scenarios, as detailed in Fig. \ref{fig5}. The earliest collected sequence is designated as the training set, while the remaining sequences are used for validation and testing. The division of the database set, training query set, validation query set, and test query set adheres to the method outlined in \cite{autoplace}. The collection scenarios are primarily residential areas with a large number of moving objects. Additionally, sequences 3-1 and 3-3 are collected during two time periods separated by a long interval. Over time, changes in object layout within the scenes and seasonal variations in vegetation significantly increase the difficulty of accurate recognition.

\begin{figure*}[h]
	\centering
	\includegraphics[width=0.95\linewidth]{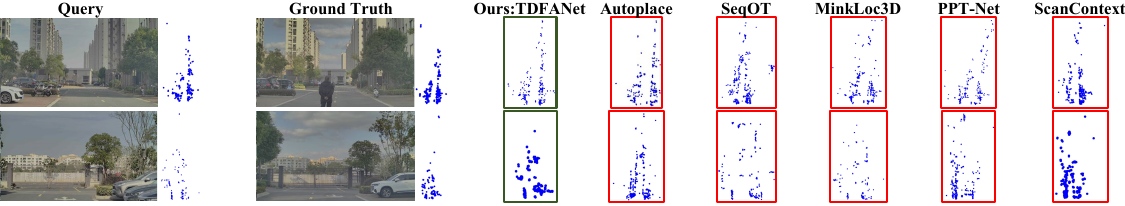}
	\centering
	\caption{Challenging query frames and reference frames retrieved by SOTA methods. Even in the presence of dynamic objects in the scene or significant appearance changes, the proposed method can accurately retrieve the top 1 reference frame, demonstrating its robustness and superiority in complex environments. Green means the retrived reference is a true positive, while red denotes false positive.}
	\label{fig6}
\end{figure*}

\subsection{Implementation Details}
For the BEV feature maps, the voxelization range is set to $[0, 69.12] \times [-39.68, 39.68] \times [-3.0, 10.0]m$ along the $(x, y, z)$ axes. The size of a voxel is set to $(0.32 \times 0.32 \times 13m)$, hence in Eq. \ref{eq5}, $h$=$w$=$0.32$. The number of frames $K$ processed by the TDFANet is set to 3. All training and evaluation experiments are performed on a single NVIDIA 3090 GPU using Pytorch 1.8.0. The Adam optimizer is employed to train the model, with an initial learning rate of 0.0008, adjusted using an exponential decay strategy with a base of 0.9. The batch size is set to 1. We set $\alpha$ and $\beta$ of the lazy quadruplet loss to 0.2 and 0.1, respectively. 4D radar scans within a radius of 5m from the query location are considered true positives, while 4D radar scans outside a radius of 10m are considered true negatives.
\subsection{Evaluation Metrics}
We follow protocols of TransLoc4D \cite{transloc4d} to evaluate our method, using $Recall@N$, which represents the percentage of correctly identified queries when given a specific number $N\in\{1,5,10\}$ of candidates.
\section{RESULTS}
\subsection{Comparison with State-Of-The-Art Methods}

Given the absence of publicly available methods for 4D radar place recognition, we adapt the latest SOTA methods for 3D LiDAR or 3D radar-based place recognition to the 4DRPR task. The comparative models include: ScanContext \cite{scancontext}, MinkLoc3D \cite{minkloc3d}, PPT-Net \cite{pptnet}, Autoplace \cite{autoplace}, and SeqOT \cite{seqot}. In addition to the four test sequences, we also present the results for all test queries after mixing different scenario databases (Seq All). The inclusion of more scenes in the database significantly increases the risk of descriptor confusion, thereby making accurate query retrieval more challenging.

As shown in Tab. \ref{table:sota}, due to the limited field of view and the sparsity of the 4D radar point cloud, ScanContext performs mediocre. MinkLoc3D and PPT-Net employ sparse convolution and multi-scale graph Transformer networks, respectively, to address the issue of LiDAR point cloud sparsity. However, these approaches are less effective on the more sparse and noisy 4D radar point clouds. In contrast, our method does not rely on a specialized network architecture for complex point cloud feature extraction. Instead, it leverages the multi-modal information from the 4D radar point cloud to design a sequential 4D radar scans aggregation method, effectively addressing the issue of point cloud sparsity. Our method consistently demonstrates superior performance across all sequences. Autoplace and SeqOT utilize LSTM and Transformer networks, respectively, to aggregate the spatio-temporal features of sequential sensor data. Compared to these methods, our method effectively mitigates the adverse effects of feature map shifts on aggregation by utilizing the velocity property of the 4D radar point cloud, and constructs a multi-scale spatio-temporal pyramid to comprehensively aggregates sequential point cloud features. Consequently, our method demonstrates superior performance in experimental evaluations.

\subsection{Ablation Study}
\setlength{\tabcolsep}{0.9mm}
\begin{table}[t]
	\caption{Ablation study of TDFANet.}	
	\centering
	\footnotesize
	\begin{center}
		\resizebox{0.96\columnwidth}{!}
		{
			\begin{tabular}{l|cccc|ccc}
				\toprule
				&  \multicolumn{4}{c|}{Components}  &\multicolumn{3}{c}{Seq All}      \\
				\cline{2-8}\noalign{\smallskip}
				
				\multirow{-2}{*}{\begin{tabular}[c]{@{}c@{}}Method \end{tabular}}
				&DPR  &FA &TSP &DA & r@1 & r@5 & r@10 \\
				\hline\hline
				\noalign{\smallskip}
				TDFANet  
				&\ding{53} &\ding{53} &\ding{53} &\ding{53} & 84.62	&91.95&93.81
				\\ 
				TDFANet-DPR  
				&\ding{51} &\ding{53} &\ding{53} &\ding{53} & 86.62\textsubscript{$\uparrow$2.00}	&92.90\textsubscript{$\uparrow$0.95} &94.82\textsubscript{$\uparrow$1.01}
				\\ 
				TDFANet-DPR-FA  
				&\ding{51} &\ding{51} &\ding{53} &\ding{53} & 88.81\textsubscript{$\uparrow$2.19}	&93.93\textsubscript{$\uparrow$1.03} &95.67\textsubscript{$\uparrow$0.85}
				\\ 
				TDFANet-DPR-FA-TSP  
				&\ding{51} &\ding{51} &\ding{51} &\ding{53} &90.11\textsubscript{$\uparrow$1.30}	&95.91\textsubscript{$\uparrow$1.98} &96.69\textsubscript{$\uparrow$1.02}
				\\ 
				TDFANet-DPR-FA-TSP-DA   
				&\ding{51} &\ding{51} &\ding{51} &\ding{51} &\bf{95.01\textsubscript{$\uparrow$4.90}}	&\bf{97.40\textsubscript{$\uparrow$1.49}} &\bf{98.18\textsubscript{$\uparrow$1.49}}
				\\ \bottomrule
			\end{tabular}
		}
	\end{center}
	\label{table:ablation}
\end{table}

To analyze the individual contribution of each component in our method, we compare the TDFANet variants that progressively apply different components. We set the plain TDFANet with all components disabled as the basic model. It only consists of BEV feature encoding and GeM pooling, aggregating sequential BEV feature maps through feature map summation. On the basis of the plain TDFANet, we use additional abbreviations to denote the application of \textbf{D}ynamic \textbf{P}oint \textbf{R}emoval (\textbf{-DPR}), \textbf{F}eature \textbf{A}lignment (\textbf{-FA}), \textbf{S}patio-\textbf{T}emporal \textbf{P}yramid (\textbf{-TSP}), and \textbf{D}eformable \textbf{A}ttention (\textbf{-DA}).

As shown in Tab. \ref{table:ablation}, the plain TDFANet establishes a decent baseline on the test sets, indicating the feasibility of using sequential 4D radar scans for 4DRPR task. Integrating dynamic point removal (TDFANet-DPR) into TDFANet significantly enhances the model's performance, highlighting the module's effectiveness in mitigating interference from dynamic and noisy points. Incorporating the feature alignment module (TDFANet-DPR-FA) results in a 2.19\% performance improvement, validating the rationale of using ego-velocity-estimated trajectories for feature alignment and the effectiveness of this module in enhancing feature aggregation. Introducing the spatio-temporal pyramid module (TDFANet-DPR-FA-TSP) and aggregating features by summing the feature maps at each pyramid level consistently improves the model's performance, highlighting the effectiveness of multi-scale spatio-temporal features in generating discriminative descriptors. Finally, incorporating deformable attention (TDFANet-DPR-FA-TSP-DA) leads to a significant performance improvement, demonstrating that this module can adaptively search for regions of interest at multiple spatial scales for aggregation, thereby effectively eliminating the impact of rotation and comprehensively aggregating features.

\subsection{Visualization}

In Fig. \ref{fig6}, we compare the retrieval results between SOTA methods and our proposed method under challenging query frames. When the scene contains dynamic objects or when the query frame exhibits significant visual differences from the ground truth frame, SOTA methods often struggle to identify the correct match. In contrast, our proposed TDFANet method effectively mitigates the impact of dynamic objects and captures robust features, successfully identifying the correct reference frame.

\section{CONCLUSIONS}
In this work, we propose the first end-to-end encoding architecture, TDFANet, that utilizes sequential 4D radar point clouds for place recognition. First, point cloud preprocessing and BEV feature encoding are introduced. On this basis, a trajectory-guided feature alignment method is proposed to align multiple BEV feature maps. Next, we design a spatio-temporal pyramid deformable feature aggregation method to extract and aggregate multi-scale spatio-temporal features. Finally, GeM pooling is used to generate the final 4D radar point cloud descriptor. Extensive experiments validate the effectiveness of TDFANet in 4DRPR tasks and demonstrate its robustness in handling dynamic objects and scene changes. 
\bibliographystyle{IEEEtran}
\bibliography{ref}
\end{document}